# Intelligent recognition of GPR road hidden defect images based on feature fusion and attention mechanism

Haotian Lv, Yuhui Zhang, Jiangbo Dai, Hanli Wu, Jiaji Wang, Dawei Wang, *Senior Member, IEEE*

*Abstract*—Ground Penetrating Radar (GPR) has emerged as a pivotal tool for non-destructive evaluation of subsurface road defects. However, conventional GPR image interpretation remains heavily reliant on subjective expertise, introducing inefficiencies and inaccuracies. This study introduces a comprehensive framework to address these limitations: (1) A DCGAN-based data augmentation strategy synthesizes high-fidelity GPR images to mitigate data scarcity while preserving defect morphology under complex backgrounds; (2) A novel Multi-modal Chain and Global Attention Network (MCGA-Net) is proposed, integrating Multi-modal Chain Feature Fusion (MCFF) for hierarchical multi-scale defect representation and Global Attention Mechanism (GAM) for context-aware feature enhancement; (3) MS COCO transfer learning fine-tunes the backbone network, accelerating convergence and improving generalization. Ablation and comparison experiments validate the framework's efficacy. MCGA-Net achieves Precision (92.8%), Recall (92.5%), and mAP@50 (95.9%). In the detection of Gaussian noise (σ=25), weak signals, and small targets, MCGA-Net maintains robustness and outperforms other models. This work establishes a new paradigm for automated GPR-based defect detection, balancing computational efficiency with high accuracy in complex subsurface environments.

*Index Terms*—Ground-penetrating radar (GPR), Hidden defects, Intelligent identification, Feature fusion, Attention mechanism

## I. INTRODUCTION

COLLAPSE accidents caused by invisible structural defects in roads have been common in recent years, with around 300 such cases reported by government agencies or authoritative media in China each year. Such accidents are still on the rise. During operational life, roads will sustain damage, such as cracks [1], loose [2], and cavities [3]. These defects are caused by various factors, such as underground pipeline issues, climate conditions, geotechnical engineering conditions, vehicle loads, etc [4]. These structural defects affect driving comfort, traffic efficiency, and safety [5]. Therefore, it is necessary to conduct regular and rapid inspections of roads to identify structural defects in advance.

The common methods for detecting hidden road defects include core drilling[6], Falling Weight Deflectometer (FWD) [7], high-density electrical method [8], Non-nuclear Density Gauge (NNDG) [9] etc. The above methods have the disadvantages of damaging road structures, high costs, low efficiency, and poor representativeness. Faced with the continuous expansion of road networks and the growing complexity of transportation systems, traditional methods of detecting hidden road defects are becoming insufficient to meet current needs. Ground Penetrating Radar (GPR) technology is founded on changes in the electromagnetic properties of various road structures. It can quickly detect internal defects, such as cracks and voids that are not visible on the surface. GPR has the advantages of high efficiency and no damage. 3D GPR can significantly improve its detection efficiency by setting up multiple pairs of electromagnetic wave transmitting and receiving antenna arrays, which is currently one of the most advanced and efficient non-destructive road measurements [10].

Hidden defects are not common in roads. It is very difficult and challenging to obtain a sufficient number of GPR images from real data for deep learning (DL). This requires a significant amount of manpower, resources, and time consumption. Many researchers have proposed techniques such as data augmentation and transfer learning to improve the quality, diversity, and scale of GPR datasets, thereby enhancing models' accuracy and generalization performance. Zhang et al. [11] proposed the incremental random sampling method, which avoids interference caused by manually captured images and improves the quality of radar images. Kim et al. [12] considered the statistical distribution of GPR data and automatically determined the upper and lower threshold amplitudes for a feature enhancement to enhance the characteristics of the target signal, thereby improving the recognition accuracy of radar images. Xiong et al. [13] improved GAN by introducing an adaptive discriminator augmentation module and a modified self attention module, effectively enhancing the

This work was supported by National Key Research and Development Program of China [grant number 2023YFB2603500]. *(Corresponding author: Dawei Wang).*

Haotian Lv, Yuhui Zhang, Jiangbo Dai, Hanli Wu, and Dawei Wang are with the School of Transportation Science and Engineering, Harbin Institute of Technology, Harbin 150090, China. (e-mail: lvhaotian@stu.hit.edu.cn; yuhuizhang@stu.hit.edu.cn; 24b932009@stu.hit.edu.cn; wuhanli@hit.edu.cn; dawei.wang@hit.edu.cn).

Jiaji Wang is with Department of Civil Engineering, University of Hong Kong, Hong Kong 999077, China. (e-mail: cewang@hku.hk).

effectiveness of GPR image generation. Qi et al. [14] added convolutional layers to Generative Adversarial Networks (GANs) to form a deep convolutional adversarial network and combined it with the Finite Difference Time Domain method to create automatically generated radar images for data augmentation. Wang et al. [15] utilized GAN with multi-scale discrimination strategy to generate GPR B-scan images generated by gprMax. Niu et al. [16] used GPR forward simulation and image enhancement methods to expand the GPR image dataset, and used YOLOv4 to recognize seven common types of underground targets in urban roads, with a detection accuracy of up to 85%. Li et al. [17] compared the performance evaluation of GPR road defect recognition using only real samples and real sample mixed gprMax simulation samples. The results showed that the model's prediction accuracy improved by 15.64%. This method effectively expanded the training dataset. Most of these data augmentation methods rely on simulation techniques to generate samples resembling real GPR images. However, these simulated data are typically based on idealized physical models and simplified environmental conditions. Consequently, these simulated data cannot fully reflect the true complexity of underground conditions. This can affect the ability of models trained on simulated data to generalize to real-world conditions.

3D GPR acquires multi-channel data to obtain complete coverage of road. Analyzing GPR images is generally dependent on the evaluator's experience, but this process is ultimately subjective and inefficient [18]. It is vital to use advanced automated approaches to improve GPR object detection effectiveness [19]. The technology of image object detection based on DL has been implemented into the recognition of GPR images [20]. Convolutional neural networks (CNN), a classic DL model, has advantages for object recognition and classification. Liu et al. [21] used a Single Shot Multibox Detector (SSD) to detect and locate the hyperbolas created by steel bars on GPR images, with a detection accuracy exceeding 90%. Zhang et al. [22] used Resnet50 to train GPR data processed by continuous wavelet transform method, achieving automatic extraction of moisture damage area. Li et al. [1] assessed YOLOv3, YOLOv4, and YOLOv5 detection performance for hidden cracks. They found YOLOv4 had the best speed and robustness while YOLOv5 had the highest accuracy. Hu et al. [23] improved the attention module to strengthen YOLOv5 acquisition of GPR images, improving the detection accuracy of voids, hyperbolic defects, and loose defects. Zhang et al. [24] improved CNN algorithms such as R-CNN, RetinaNet, YOLOv3, and YOLOv5 using Swin Transformer. They enabled the real-time detection of voids and achieved the best detection accuracy using Swin-YOLOv5. Yang et al. [25] applied two DL models, CP-YOLOX and SViT, to 3D GPR data, achieving high-precision identification of cracks, voids, poor interlayer bonding, and mixture segregation. Liu et al. [26] used a novel FeMViT network for GPR image recognition, and the recognition performance was significantly higher than that of the original Transformer. Liu et al. [27] introduced MobileNetV2 and Convolutional Block Attention Module (CBAM) to improve the YOLOv4 network, achieving better GPR image recognition performance than YOLOv5. With the emergence of the YOLOv8 model, research on its application to internal road defect detection and structure recognition has gradually developed. Guo et al. [28] used Faster R-CNN, YOLOv5, and YOLOv8 models to recognize underground radar images inside roads intelligently, and the results showed that YOLOv8 is the state-of-the-art model. Wang et al. [29] developed a YOLOv8 model that includes CBAM and a Simple Linear Iterative Clustering Phash method, which improves the efficiency of intelligent detection of underground defects. Liu et al. [30] introduced the Selective Kernel Networks attention mechanism to improve the YOLOv8 model, further increasing the sensitivity range and improving the performance of internal crack detection in pavement structures.

The above methods often use the classic single-stage algorithm - YOLO series algorithm, which has the advantages of high speed, high accuracy, easy deployment, and high update frequency, and is widely used in computer vision. Among them, YOLOv8 version is relatively advanced. By improving the network architecture, loss function, feature fusion mechanism, etc., it can better cope with complex real-world scenarios and improve the applicability of the model in various application fields. However, YOLOv8 and its improved models have not yet performed better in identifying internal road defects. These research algorithms mainly focus on extracting local features from the convolutional layers of DL models, but these methods often overlook the correlations and global structural information between features at different scales. As a result, in complex underground environments, the model may not be able to fully capture the overall shape or important details of the target, limiting recognition accuracy. In addition, although some studies have started to introduce attention mechanisms such as CBAM [23, 27, 29] and self-attention [25] to improve the global perception ability of the model, the application of global attention in GPR image processing is still insufficient. Most studies still focus mainly on local features and fail to effectively combine global and local information, resulting in the unsatisfactory performance of models when dealing with complex backgrounds or data with significant interference.

Therefore, this article aims to study a multi-category road hidden defect recognition framework that improves the quality of feature extraction and cross-dimensional feature interaction. A large number of images of internal road defects were collected and verified on-site. On this basis, a dataset containing 1800 images of them was constructed using the DCGAN model for data augmentation. Subsequently, an improved YOLOv8 model combining Multi-modal Chain Feature Fusion (MCFF) and Global Attention Mechanism (GAM) modules was proposed to enhance the correlation and global structural information between features of different scales, thereby improving the accuracy of small object detection in GPR images. At the same time, the introduction of transfer learning methods further improves the detection performance.

## II. GPR DATA COLLECTION METHODS

### A. 3D GPR System

The high-frequency electromagnetic waves emitted by GPR form reflected electromagnetic waves in materials with different dielectric constants, thus providing information about the road structure. The reflection coefficient of the electromagnetic wave is given by the following (1) [31].

$$R = \frac{\sqrt{\varepsilon_1} - \sqrt{\varepsilon_2}}{\sqrt{\varepsilon_1} + \sqrt{\varepsilon_2}} \quad (1)$$

where R is the reflection coefficient; $\varepsilon_1$ and $\varepsilon_2$ are the relative dielectric constants of the upper- and lower-layer materials. The greater the difference between $\varepsilon_1$ and $\varepsilon_2$, the greater the amplitude of the reflected wave. When defects inside the road structure are present, they will cause changes in the relative dielectric constant of the material, thereby displaying specific amplitude variation patterns on radar images.

The 3D GPR collects signals through a multi-channel antenna array, records electromagnetic wave data reflected by underground targets, and generates 3D spatial imaging through data integration to visually display the spatial distribution of targets; The 2D GPR only collects data through a single survey line. Compared to 2D GPR, 3D GPR has higher detection efficiency and can achieve full spatial coverage.

The antenna in 3D GPR can be divided into air-coupled and ground-coupled antennas. The air-coupled antenna is placed at a certain height from the road surface and can collect reflected high-frequency electromagnetic signals. The ground-coupled antenna usually is placed less than 5 cm from the road surface. The incident wave signal and the road surface reflection signal will produce signal coupling overlap at the signal receiver, and it is difficult to determine the air layer in the GPR image for the ground-coupled antenna. The characteristics of them are shown in Table I.

TABLE I
ANTENNA CHARACTERISTICS OF DIFFERENT COUPLING MODES

| Index | Air-coupled antenna | Ground-coupled antenna |
|---|---|---|
| Height from road surface (cm) | 15-80 | <5 |
| Driving speed (km/h) | 20-80 | <20 |
| Frequency range (MHz) | 500-3000 | 100-1500 |
| Detection depth (m) | 0-2 | 0-10 |
| Detection purpose | Layer thickness Density Internal crack Debonding Moisture | Moisture Subsidence Sinkholes Internal crack Underground utilities (pipes, cavities, etc.) |

### B. Original Dataset construction

The GeoScope 3D GPR system was used to detect hidden defects in the road. The system mainly comprises GeoScope™ MKIV radar, DXV1808 ground-coupled antenna, distance measurement indicator (DMI), and real-time kinematic (RTK). The application of stepped frequency technology in antennas can actively consider shallow high-resolution imaging combined with deep structural analysis, meeting the requirements of high-quality GPR image generation for different defect distribution depths [32]. The data collection site is the urban streets in Harbin, China, as shown in Fig. 1.

The specific details are as follows: ① The road is an asphalt pavement and a main urban road with a busy traffic environment. Harbin is located in northern China, with a general terrain trend of high in the south, low in the north, and the lowest river in the middle, with an altitude of 114-170m. The temperature zone is a seasonal freezing zone. The road is prone to cracks, uneven settlement, and void caused by freeze-thaw cycles. ② Collect data during the period of least traffic volume from 0:00 to 6:00 in the morning to ensure smooth traffic and reduce environmental interference to ensure that the quantity and quality of GPR data meet the requirements for building an intelligent detection dataset, the maximum speed is set at 15 kilometers per hour, the length of the data collection road is 78.5 kilometers, and the total length of the GPR survey line exceeds 500 kilometers. The main technical and data acquisition parameters of the 3D GPR system are listed in Table II. Among them, transverse sampling interval represents the distance interval between each data collection along the driving direction. Time window represents the duration of each data collection by the receiving antenna. Dwell time represents the time for each frequency to emit electromagnetic waves. The above three parameters will determine the maximum acquisition speed of GPR antenna. The 3D GPR system, configured with a frequency range of 90–1000 MHz and a 7.65 cm transverse sampling interval, combines stepped frequency technology to ensure sufficient resolution for detecting cracks, cavities, and concave.

TABLE II
MAIN TECHNICAL PARAMETERS AND DATA ACQUISITION PARAMETERS OF GPR SYSTEM

| Parameter | Value |
|---|---|
| Frequency range | 90-1000 MHz |
| Number of channels | 8 |
| Channel spacing | 164 mm |
| Effective acquisition width | 1.32 m |
| Transverse sampling interval | 76.5 mm |
| Time window | 125 ns |
| Dwell time | 5.000 μs |

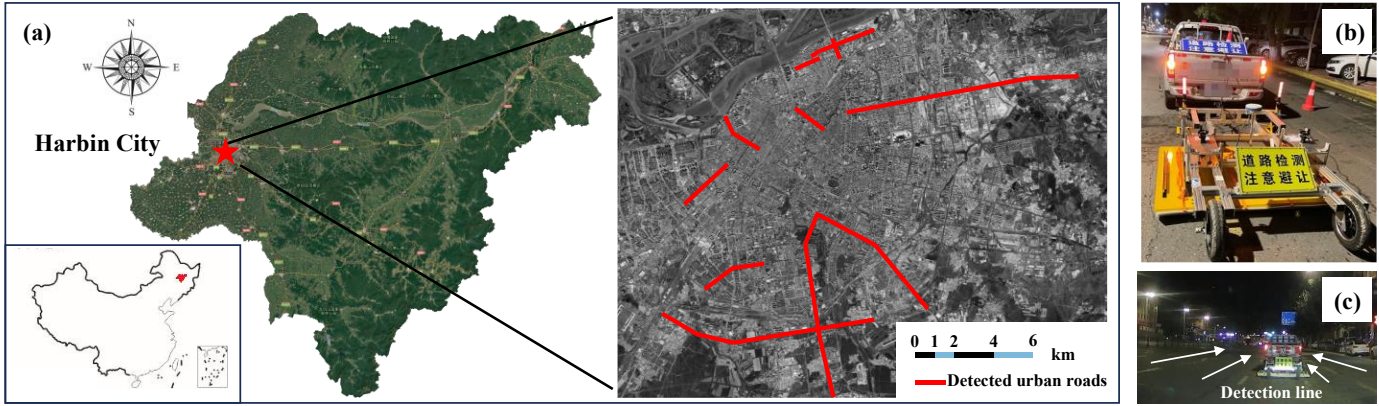

**Fig. 1.** Harbin City 3D GPR detection. (a) Location and urban road of Harbin City; (b) On-site detection with GPR; (c) Schematic diagram of the GPR detection line

The establishment of the dataset includes four steps: acquisition, filtering, capturing, and labeling. Once the GPR data has been acquired, the Examiner software can be used to obtain feature information in 3D, namely the vertical, transverse and horizontal sections, as shown in Fig. 2.

In order to obtain clear radar images and better construct datasets, data preprocessing must be performed on the original GPR data. The Inverse Selective Discrete Fourier Transform (ISDFT) [32] converts data from the frequency domain to the time domain for further analysis and interpretation. Interference suppression is employed to identify and mitigate interference from surrounding vehicles, mobile communication base stations, and other environmental sources during GPR data collection. This process utilizes advanced signal processing techniques to reduce dominant interference components while preserving critical subsurface features[33]. Gradual low-pass and background removal filters are used to eliminate interference caused by high-power shallow reflection signals. The specific GPR data preprocessing parameter settings are listed in Table III.

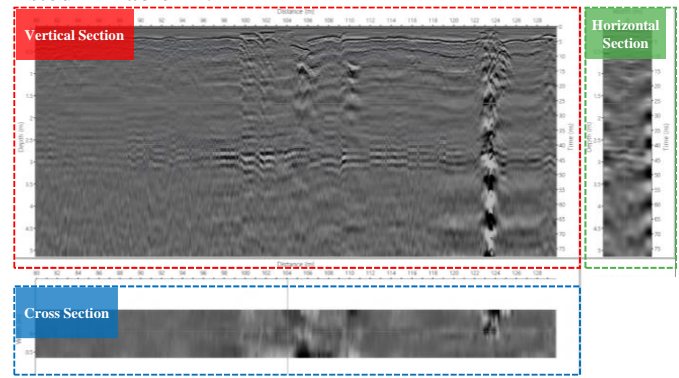

**Fig. 2.** Vertical, cross, and horizontal processed images of the GPR data

TABLE III
GPR DATA PREPROCESSING PARAMETER SETTINGS

| Interference suppression | Parameter | ISDFT | Parameter | Gradual low-pass filter | Parameter | Background removal filter | Parameter |
|---|---|---|---|---|---|---|---|
| Power Limit | 16 | Attenuation | 0.035 | Vertical Filtering | Enabled | Filter Length | 10 m |
| Output Percentages | Enabled | Window type | Tukey | Horizontal Filtering | Disabled | Removal | 100 % |
|  |  | Tukey alpha | 0.35 |  |  | Start Depth | 0 ns |
|  |  | Use full BW | Enabled |  |  | Transition Zone | 1 ns |
|  |  | Maximum frequency | 1000 MHz |  |  | Filter Mode | Sliding Window Mean |
|  |  | Cut-off limit | 90 MHz |  |  |  |  |

The GPR images can be obtained after image filtering, and the defection can be recognized due to the dielectric property variation of different materials. The images with different characteristics can be thereafter used to recognize different hidden defections. The hidden defect images include cavities, concaves and cracks. The characteristics of them are shown in Table IV.

The captured GPR images can be marked by labeling software. The label information is stored as an XML file for subsequent model training.

TABLE IV
CLASSIFICATION OF HIDDEN DEFECTS

| GPR image | Hidden defect | Image interpretation |
| --- | --- | --- |
| 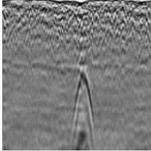 | Cavity | it is usually a hyperbolic shape with a large span opening downward, and diffraction / multiple waves are more obvious. |
| 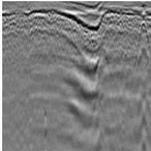 | Concave | The horizon line has a strong reflection area; the amplitude is enhanced; the waveform is disordered; the sinking of the event is obvious; the diffracted wave and multiples are obvious. |
| 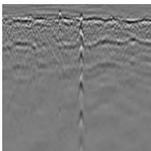 | Crack | High overall amplitude, locally convex waveform, and well-defined diffracted wave at the fracture boundary. |

*C. Data validation*

In order to further accurately locate the hidden defect area and determine its impact scale. After using 3D GPR to identify hidden defects, other methods need to be used for retesting verification. The purpose of retesting verification is to collect the condition of various structural layers of the road through excavation, drilling and other technological means, to provide detailed and scientific basis for road repair work.

The specific process of retesting verification is as follows: ① At the location of hidden defects detected by 3D GPR, a 2D GPR is used for grid based encrypted wire harness detection. ② Real time marking of the starting position of abnormal signals on the road, accurately determining the location and area of hidden defects. Drill or core at the geometric center of the marked concealed area. Use an endoscope to explore real internal images and verify the accuracy and reliability of 3D GPR detection results. Based on the characteristics of GPR images, retesting verification was conducted on the original samples representing the top 30% in terms of severity of hidden defect development. The retest verification process is shown in the Fig. 3.

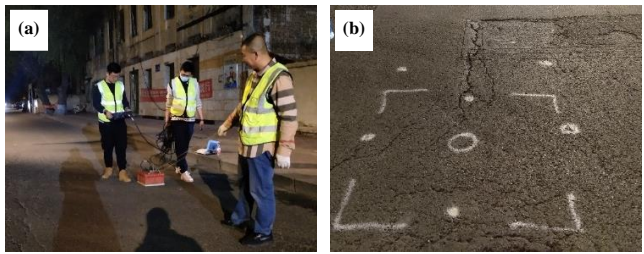

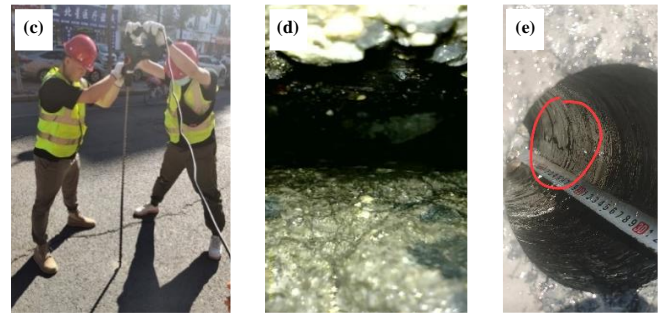

**Fig. 3.** Hidden defects retest verification. (a) 2D GPR Encryption line detection (b) Mark the location and area of hidden defect (c) Drilling at defect locations (d) Endoscopic inspection of cavity locations (e) Core inspection at crack location

### III. MULTI-CATEGORY ROAD HIDDEN DEFECT RECOGNITION FRAMEWORK

We have collected high-quality GPR images for on-site defect detection in Section 2. However, the scarcity of real data samples is an important factor that hinders the improvement of accuracy in identifying hidden defects. In addition, even though the YOLO series has been updated to YOLOv8, it has not yet shown better performance in identifying internal road defects. Therefore, we propose a performance improvement framework for multi-class road hidden defect recognition. Its process is mainly divided into 5 stages:

1. Firstly, use DCGAN network to generate new GPR images and increase the number of datasets. Balance the number of GPR images for different categories of defects to achieve data augmentation.

2. Create a MCFF module to effectively extract multi-scale features from images, improve the quality of feature extraction, and enhance the performance of hidden defect recognition.

3. Utilize GAM to enhance the ability of DL networks to extract key features from images, thereby further improving the accuracy of hidden defect recognition.

4. Based on the YOLOv8 image recognition network, effectively integrate MCFF and CAM modules, fully utilize their performance, and establish a DL algorithm MCGA-Net for multi-class road hidden defect recognition.

5. Combine transfer learning ideas. The weight files obtained from pre-training MCGA-Net using the MS COCO dataset are applied to the training of the GPR dataset to further improve network performance.

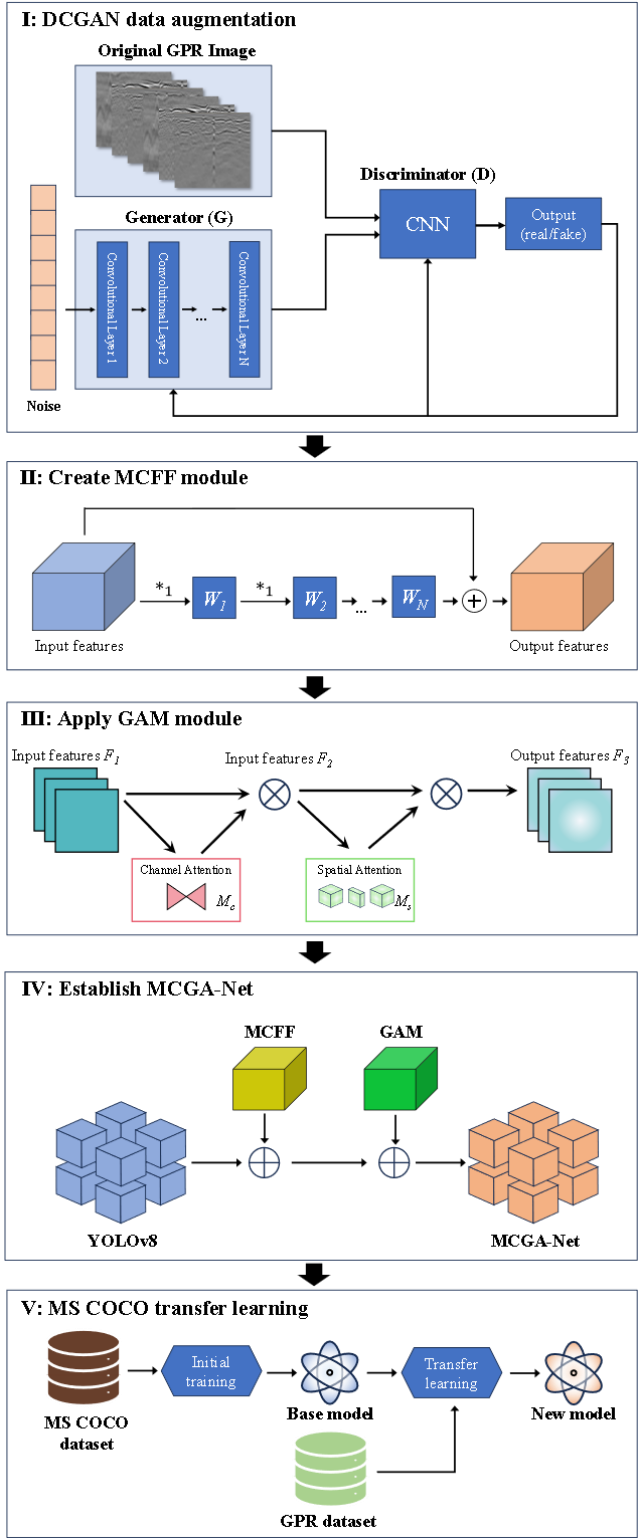

**Fig. 4.** Multi-category road hidden defect recognition framework

The advantages of the above multi-category road hidden defect detection performance improvement methods are as follows. Firstly, DCGAN can generate synthetic images that are highly similar to real GPR images, capturing the complex patterns and structures of GPR data. These generated images can be effectively used for data augmentation, compensating for the shortcomings of real data, increasing the diversity of training data, and thereby improving its generalization ability in different scenarios.

Secondly, varying characteristics and scale of the defects appear in the GPR image. By fusing different modalities, the MCFF has distinguished GPR underground targets better from background noise, improving capability to discriminate weak signals or small targets. The GAM established relationships among the features across the full span of the image dimensions, improving the ability understand the image global structure. This will be beneficial for identifying specific targets of different sizes in GPR data. With these two modules anticipating detecting features differently and at different scales, MCGA-Net augments attention to important features. This mechanism makes the model adaptable to complex underground environments and improves its generalization capability in different scenarios.

Thirdly, the GPR image large-scale data is difficult to acquire while MS COCO dataset has a universal large scale data set containing objects with various categories or scene. By utilizing pre-trained weights on COCO, the need for a large amount of GPR data can be reduced, significantly accelerating convergence speed, reducing training time, and improving model performance under limited data conditions. The proposed framework was implemented using PyTorch, an open-source DL library, with GPU acceleration to optimize training efficiency. The details of each stage are shown in Fig. 4.

*A. DCGAN Data augmentation*

Firstly, Considering the small number of GPR image samples, this paper uses GANs [34] to expand the dataset and improve the training accuracy of the model. Using GAN as feature extractors can learn useful feature representations from a large amount of unlabeled data and then apply these features to supervised learning. However, the training of GANs is unstable and sometimes produces strange results. DCGAN [35] combines CNN with GAN, as shown in Fig. 5. The generator and discriminator of DCGAN have replaced the fully connected network of the original GAN with CNN. By using stride convolutions instead of deterministic spatial pooling functions and global average pooling instead of fully connected layers, the stability of the model has been improved. Using BN batch normalization on the output layer of the generator (G) and the discriminator (D) input layer can help solve the poor initialization problem and facilitate gradient flow toward deeper networks.

To further combat mode collapse under sparse GPR data conditions, our DCGAN integrated two synergistic strategies: ① Label smoothing on discriminator targets (real=0.9, fake=0.1), which mitigated gradient saturation by preventing excessive confidence in real/fake classification; ② A stochastic augmentation module that applied horizontal flips and controlled Gaussian noise ($\sigma = 0.05$) to input batches, perturbing local features while preserving global GPR waveform integrity.

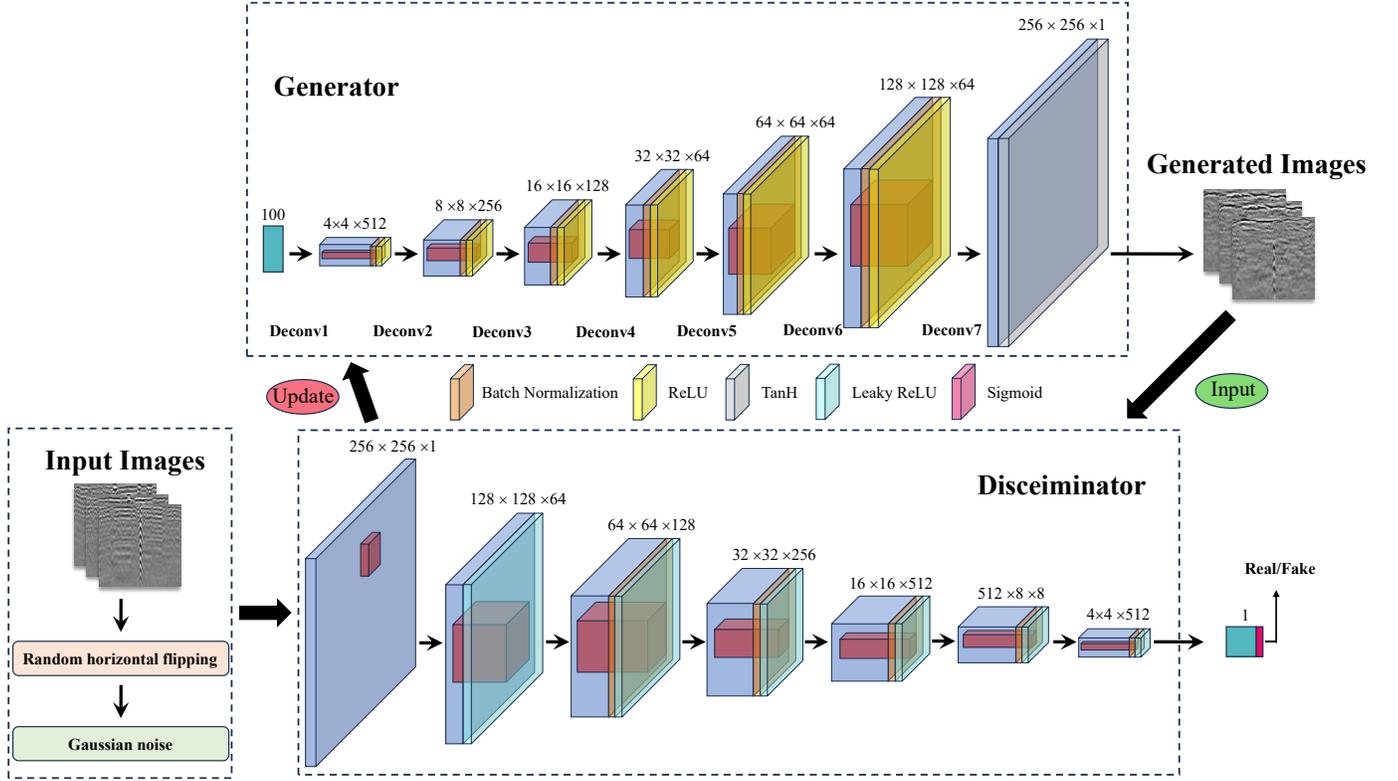

**Fig. 5.** Schematic diagram of DCGAN structure

## B. Create MCFF model

Traditional feature fusion methods only consider the structural information of data features, which reduces the quality of feature extraction from data. In order to improve the quality of features, the generalization of features is enhanced by considering the interaction between various dimensions of input data. Assuming that the input data $X \in R^{I \times J \times K}$ of the L layer, the quality of feature extraction from the input data is improved by multi-step chain integration of the information of modules I, J, and K, and the density of structural details of the features is increased. The mathematical expression of the MCFF method is shown in (2), as illustrated in Fig. 6:

$$Y = X *_1 W_1 *_1 W_2 *_1 W_3 \qquad (2)$$

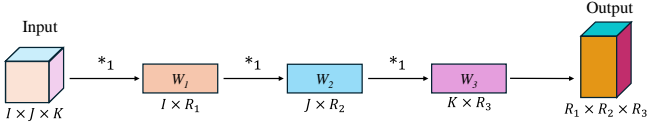

**Fig. 6.** Multi-modal chain feature fusion (MCFF)

Among them, the single-mode trainable weight matrix $W_1 \in R^{I \times R_1}$, $W_2 \in R^{J \times R_2}$, and $W_3 \in R^{K \times R_3}$, output tensor $Y \in R^{R_1 \times R_2 \times R_3}$. $*_N$ is the Einstein product [36], which is a contracted product used to represent efficient operations between tensors and matrices. When N=1, the Einstein product is a standard matrix multiplication. By performing multi-step single-mode Einstein multiplication operations, the quality of feature extraction can be effectively improved, and the structural information density of features can be increased.

When $R_1$=I, $R_2$=J, $R_3$=K, performing feature extraction operations of the same order and dimension on the input data can introduce coarse-grained feature information by adding residual connections. At this point, the mathematical expression of the MCFF is shown in (3), as illustrated in Fig. 7:

$$Y = X + X *_1 W_1 *_1 W_2 *_1 W_3 \qquad (3)$$

**Fig. 7.** Same order and dimension MCFF

Among them, the single-mode trainable weight matrices $W_1 \in R^{I \times I}$, $W_2 \in R^{J \times J}$, and $W_3 \in R^{K \times K}$, output tensor $Y \in R^{I \times J \times K}$. (2) integrates coarse-grained and fine-grained information from the input data, preserving the original information while enhancing the extraction of fine-grained details through feature extraction operations. This approach not only improves the generalization capabilities of the feature extraction process but also enhances the robustness of the model when dealing with noisy or incomplete data. Residual connections can effectively alleviate the gradient vanishing problem in deep neural networks and improve training efficiency.

## C. Apply GAM model

GAM is a technique used to enhance the performance of deep neural networks, particularly in computer vision tasks. The technical principle of this mechanism is to enhance cross-dimensional interaction by preserving channel and spatial information. Compared to previous methods such as Senet [37] and CBAM [38], the GAM focuses on maintaining the importance of global interaction when processing channel and spatial information.

The GAM combines 3D permutation and multi-layer perceptron for channel attention and includes a convolutional spatial attention submodule. These components reduce information loss and amplify global dimensional interaction features.

Channel attention maps and spatial attention maps are used to capture essential features in the channel and spatial dimensions, respectively. The channel attention submodule uses a three-dimensional arrangement to preserve three-dimensional information. A two-layer multi-layer perceptron (MLP) is then used to enhance the cross-dimensional channel space dependencies. In the spatial attention submodule, two convolutional layers are used for spatial information fusion to focus on spatial information. The same channel attention submodule reduction rate r is used as the Bottleneck attention module (BAM). Meanwhile, the maximum pooling operation reduces information and generates negative contributions. The pooling operation has been removed from this module to preserve the feature mapping further. These attention maps interact with the input feature maps through element-level multiplication operations, achieving global cross-dimensional information exchange. The structure is shown in Fig. 8.

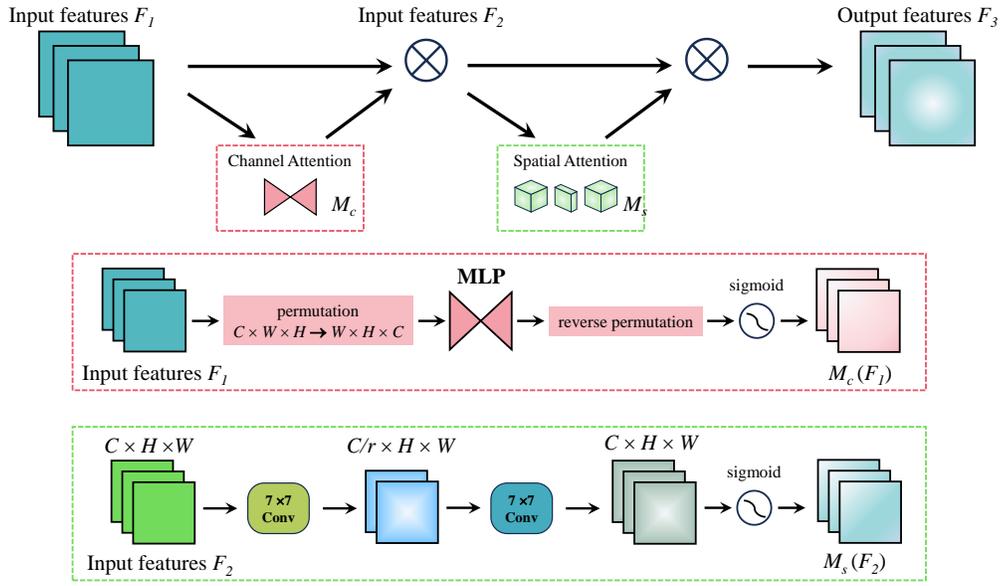

**Fig. 8.** Global Attention Mechanism (GAM) structure

## D. Establish MCGA-Net

You only look once (YOLO) target detector is mainly composed of input, backbone, neck, and head [39]. YOLOv8 is the latest version of the YOLO series, which improves network architecture, introduces advanced feature fusion and optimization strategies, and further enhances detection accuracy and robustness while maintaining efficient real-time detection. MCGA-Net is an upgrade and improvement of the YOLOv8 model, aimed at further enhancing the accuracy of GPR image detection and recognition. Early-stage features captured at position 3 predominantly encode localized texture patterns but exhibit limited global contextual relationships. The channel-spatial attention mechanism within GAM strategically amplifies cross-region dependencies during the critical transition from low-level to mid-level semantics. By dynamically weighting inter-channel correlations and spatial feature interactions, this mechanism establishes contextual linkages between defect signatures and their surrounding subsurface structural patterns, thereby enhancing feature discriminability in complex underground environments. In addition, MCFF module is integrated into the neck architecture at position 21 to hierarchically aggregate multi-scale defect signatures. This module enables cross-resolution feature fusion, capturing defects textures and geometries while suppressing noise. Such hierarchical integration ensures robust representation of defect morphologies across diverse subsurface conditions, thereby enhancing the model's generalization capability under varying scenarios. The overall structure of the model is shown in Fig. 9.

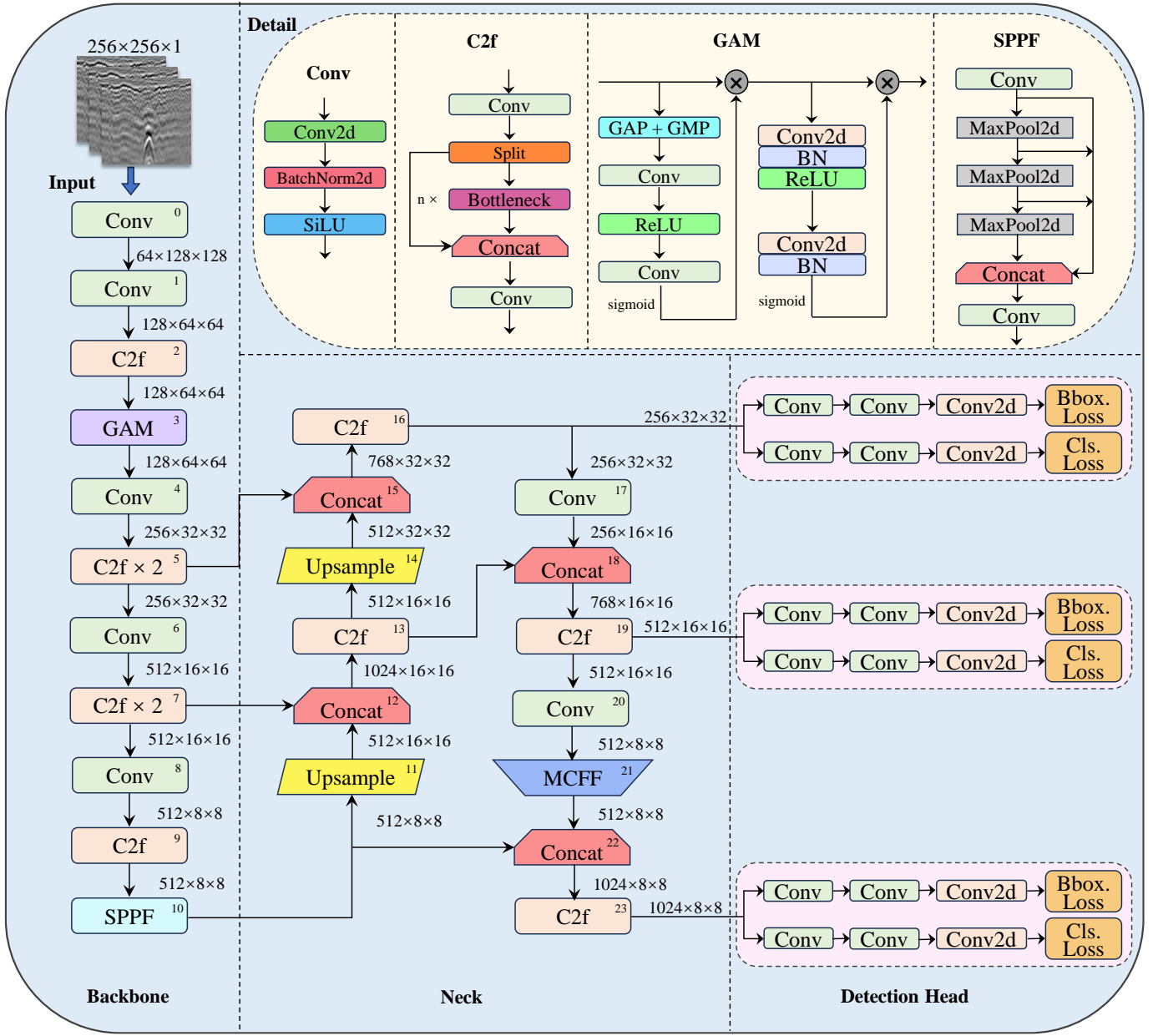

**Fig. 9.** MCGA-Net structure

*E. MS COCO transfer learning*

In the domain of DL, transfer learning is an important technique. It improves efficiency and performance by utilizing weight parameters of pre-trained models for a new task. Transfer learning is particularly valuable for situations with limited computing resources or sample size. The main idea of transfer learning is to pre-train a model on a large dataset, and use the learned weights as initialization for training it on new tasks or datasets so that fine-tuning takes much less time and data than from scratch.

In the field of image processing, the MS COCO (Common Objects in Context) dataset [40] is widely used. It includes 80 object categories and over 200,000 annotated images, making its pre-trained weights an ideal choice for transfer learning. Specifically, models pre-trained on MS COCO have learned many useful features representations that can teach and generalize the use of GPR images classification. Due to limited number of real samples, the specific transfer learning used within this study is as follows:

To begin with, the enhanced MCGA-Net model is trained for 50 epochs on the MS COCO dataset. Its weight parameters are preserved. Subsequently, the weights are loaded as initial parameters for the new assignment. By fine-tuning with the new dataset, that the general features learned by the pretrained model improved the convergence speed of training on the new task, while also improving the model's generalization ability.

## IV. EXPERIMENTAL TRAINING CONFIGURATION

All experiments were conducted on a high-performance computing platform equipped with an NVIDIA RTX 3060 GPU, utilizing PyTorch for model training and evaluation and Adam optimization. When training DCGAN, the initial learning rate is 0.0001. When conducting MCGA Net training, the initial learning rate is 0.01. These configurations balanced computational efficiency and model convergence.

The GPR images collected in real-world environments contain complex backgrounds such as road surface textures, random noise, and environmental electromagnetic interference, all of which can potentially interfere with the accuracy of defect identification. The number of training datasets is shown in Table V, and the original image dataset is referred to as GPR-Ori. The dataset expanded using DCGAN is referred to as GPR-Aug. The quantity of each dataset is shown in the table below. DCGAN effectively solves the problem of class imbalance, alleviates overfitting, and improves the accuracy of the classifier. Its effectiveness was subsequently validated during the supervised classification process.

TABLE V
DATASETS OF THE GPR IMAGES

| Category | Cavity | Concave | Crack | Total |
|---|---|---|---|---|
| Original amount | 496 | 214 | 260 | 970 |
| Test | 136 | 66 | 89 | 291 |
| GPR-Ori (Train) | 360 | 148 | 171 | 679 |
| GPR-Aug (Train) | 464 | 534 | 511 | 1509 |

To rigorously evaluate the image generation capability of DCGAN, we employ two complementary metrics: Fréchet Inception Distance (FID) and Energy gradient. FID serves as a principled measure of distributional divergence between synthetic and real data manifolds, while the Energy Score quantifies geometric diversity in generated samples.

The FID metric computes the minimum Wasserstein-2 distance between multivariate Gaussian approximations of feature distributions derived from real and generated images. It can better describe the difference between two distributions. When calculating FID, first extract a set of samples from the real data distribution and the generated model, and then use a pre trained Inception network to extract feature vectors from these samples. Next, calculate the mean and covariance matrix of the two distributions, and calculate the Fréchet distance between them to obtain the FID value. The smaller the FID value, the closer the image generated by the generative model is to the true data distribution. The FID is calculated as:

$$\text{FID}^2 = \|\mu_1 - \mu_2\|^2 + \text{Tr}(\Sigma_1 + \Sigma_2 - 2(\Sigma_1 \Sigma_2)^{1/2}) \quad (4)$$

Where $\mu_1$ and $\mu_2$ represent the mean vectors of real data and generative models, $\Sigma_1$ and $\Sigma_2$ represent the covariance matrices of the two, and Tr represents the trace of the matrix.

Energy gradient is an indicator used to measure image clarity. By calculating the image gradient's total energy, the image's edge intensity and detail clarity can be evaluated. Usually, the more edges and details in an image, the higher the Energy gradient value, and the clearer the image.

$$I(f) = \sum_y \sum_x (|f(x+1,y) - f(x,y)|^2 + |f(x,y+1) - f(x,y)|^2) \quad (5)$$

f (x, y) is the gray value of the pixel at position (x, y) in the image f.

In this paper, Precision (P), Recall (R), mean Average Precision (mAP) are used to evaluate the detection effect of the model.

$$P = \frac{TP}{TP + FP} \quad (6)$$

$$R = \frac{TP}{TP + FN} \quad (7)$$

Where TP is the number of the defections correctly detected, FP is the number of the non-defections treated as defection, and FN is the number of the defections treated as non-defection.

P and R are two related quantities. The area enclosed by the curve and coordinate axis formed by combining these points is Average Precision (AP), which can be regarded as the average of the accuracy values corresponding to different recall rates. mAP represents the average Precision of four defects. They are calculated using the following equations:

$$AP = \int_0^1 P(r) dr \quad (8)$$

$$mAP = \frac{\sum_{n=1}^{N} AP(n)}{N} \quad (9)$$

Where P(r) represents the curve corresponding to the P and R, and N represents the number of defects.

## V. RESULTS

### A. DCGAN data quality

The training progression of the DCGAN, as illustrated in Fig. 10, reveals a series of distinct observations across various epochs. Initially, after 500 epochs, the generated images showed minimal presence of abnormal features; however, the images were characterized by a high level of background noise. As training continued to 1000 epochs, image features began to emerge, yet noise artifacts remained prominent. Upon reaching 2000 epochs, while some contours of abnormal features started to become discernible, significant noise persisted, compromising the overall image quality. By 3000 epochs, the features within the images appeared inconsistent and unstable, indicating that the network was still struggling to learn a coherent representation. Only at epoch 4000 was a well-defined and stable feature achieved with the DCGAN samples. By this point, the generated images were significantly higher in quality than at any previous stage of training.

To evaluate the generative performance of DCGAN, we implemented a stratified random sampling of 200 images from both original GPR field data and synthetically generated outputs across progressive training intervals. Quantitative metrics, including FID and Energy gradient, were computed to holistically assess distributional alignment and defect signature preservation.

As detailed in Table VI, the FID demonstrates a statistically significant reduction from 260.81 to 31.26 over increasing training epochs, confirming progressive convergence toward the real data manifold. Concurrently, Energy gradient measurements reveal sustained high complexity, indicating DCGAN's capacity to augment limited field datasets while preserving textural heterogeneity essential for discriminating diverse defect typologies.

In addition to these objective metrics, visual inspections of the images corroborated the quantitative findings. The generated images were clearer and better contrast qualities ideal for underground defect detection applications.

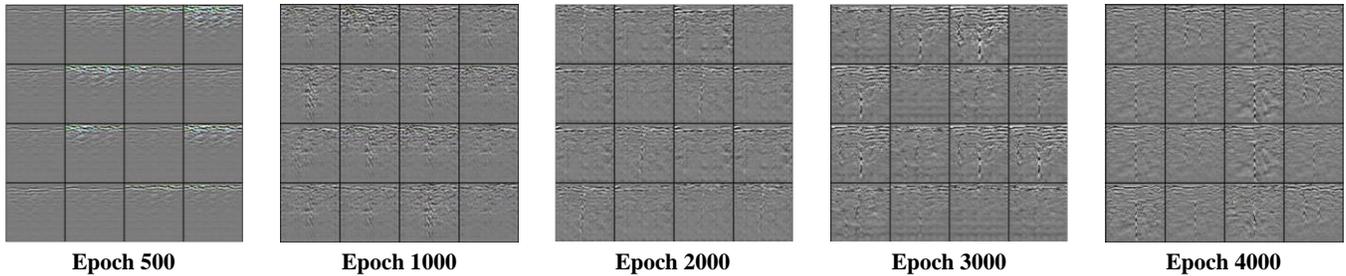

**Fig. 10.** Images generated by DCGAN

TABLE VI
THE RESULTS OF IMAGE QUALITY EVALUATION INDEXES

| Image | Epoch | FID | Energy |
|---|---|---|---|
| Original image | - | - | 7650229087 |
| DCGAN image | 1000 | 260.81 | 3545483625 |
| | 2000 | 114.59 | 4685775102 |
| | 3000 | 64.32 | 6864259736 |
| | 4000 | 31.26 | 8059277021 |

*B. Detection evaluation*

To validate the effectiveness of the proposed algorithm, a comparative analysis was conducted across four different scenarios: (1) YOLOv8 using the GPR-Ori dataset, (2) YOLOv8 using the GPR-Aug dataset, (3) MCGA-Net using the GPR-Aug dataset, and (4) MCGA-Net using the GPR-Aug dataset in conjunction with MS COCO pre-trained weights. Fig. 11 presents the training loss curve for the algorithm under investigation. The algorithm presented in this study is supervised by the box_loss, cls_loss and dfl_loss functions. Box_loss is exclusively dedicated to the regression of bounding boxes. The objective of this component is to quantify the discrepancy between the actual and predicted positions of bounding boxes. Cls_loss is to assess the model's ability to analyze and categorize objects. Dfl_loss represents the prediction accuracy of the bounding box position, contributing to refining the model's spatial predictions. MCGA-Net exhibited rapid convergence, with training and validation losses stabilizing significantly after 40 epochs. After epoch 160, both training and validation loss values stabilized below 1.15, indicating robust convergence. To verify this observation, the training was extended to 200 stages. The total loss was reduced by less than 0.1, with no material impact on model performance or conclusions. Notably, MCGA-Net with MS COCO pre-trained weights consistently achieved the lowest loss curve across all configurations.

Table VII provides a comprehensive comparison of the performance indicators across the four experimental scenarios. As evidenced by the results, each modification markedly enhanced the efficacy of the YOLO v8 model. MCGA-Net algorithm had led to a notable enhancement in Precision, Recall and mAP@50. DCGAN improved the accuracy and robustness of defect detection from the perspective of increasing image diversity and quantity. Furthermore, the utilization of pre-trained weights from the MS COCO dataset not only accelerated the training process but also provided a model with greater generalizability for GPR datasets.

Fig. 12 (a) illustrates the average Precision - Recall curve for the studied algorithm. The algorithm achieved mAP@50 of 0.967 across all categories. The high Precision scores demonstrated the model's efficacy in detecting diverse defect types, exhibiting both robustness and accuracy. This provided further evidence of the practical potential of MCGA-Net algorithm as an automatic defect locator.

Furthermore, Fig. 12(b) presents the confusion matrix for the algorithm, which provides a detailed evaluation of the model's classification performance. The matrix demonstrated that the model exhibits consistent accuracy across all classes. It evidenced robustness and accuracy in detecting hidden defects.

In conclusion, the proposed algorithm not only achieved lower loss values and faster convergence during training but also exceled in detection accuracy. These results confirmed the applicability of this model in practical defect detection as an innovative and highly efficient method.

TABLE VII
COMPARISON EXPERIMENTS

| Experimental scenarios | Precision (%) | Recall (%) | mAP@50 (%) |
| --- | --- | --- | --- |
| YOLOv8 GPR-Ori | 75.4 | 76.9 | 80.9 |
| YOLOv8 GPR-Aug | 89.1 | 89.6 | 92.1 |
| MCGA-Net GPR-Aug | 92.8 | 92.8 | 95.9 |
| MCGA-Net + COCO pretrain GPR-Aug | **92.8** | **94.0** | **96.7** |

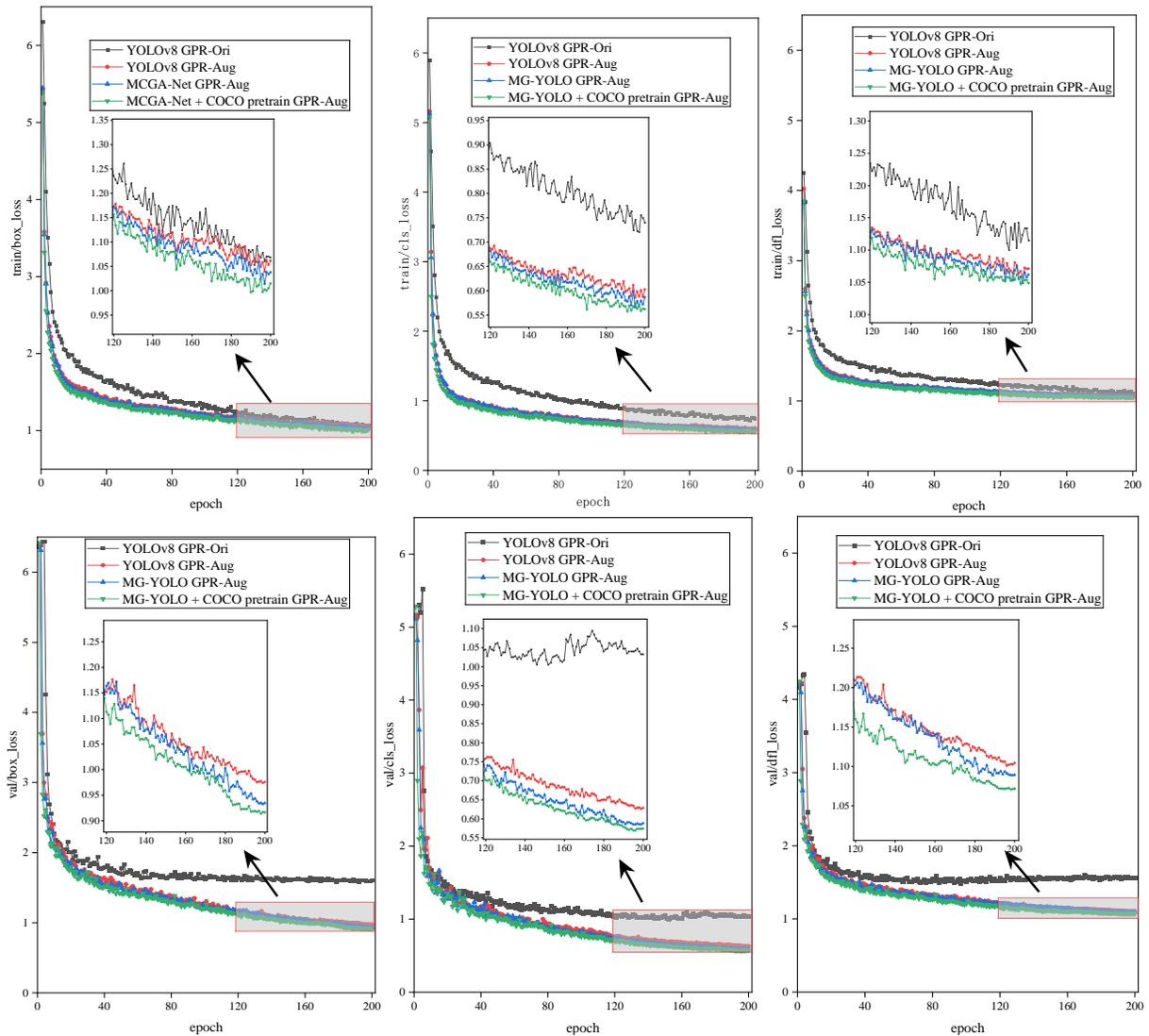

**Fig. 11.** Model training evaluation. The various loss curve values of MCGA-Net + COCO pretrain GPR-Aug are the lowest

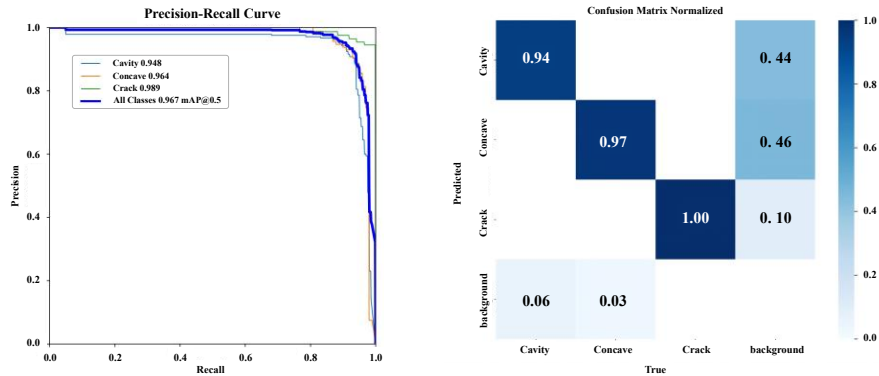

**Fig. 12.** (a) Precision-Recall Curve; (b) Confusion Matrix

*C. Ablation experiments*

The objective of the ablation study is to conduct a quantitative evaluation of the impact of each enhancement step on the overall performance of the algorithm. This approach facilitates the comprehension of the contributions of each component to the algorithm, thereby elucidating the efficacy of them in enhancing object detection accuracy.

The results of these experiments are summarized in Table VIII. The original YOLOv8 algorithm initially demonstrated a precision of 75.4%, recall of 76.9%, and mAP@50 of 80.9%. However, the incorporation of the augmented GPR dataset and the integration of the MCFF and GAM modules contributed to a notable enhancement in the algorithm's performance.

Initiating with DCGAN-based synthetic data generation, the framework addressed inherent data scarcity through adversarial training. Ablation experiments demonstrated DCGAN's efficacy in improving detection metrics by +13.7% Precision, +15.6% Recall, and +13.9% mAP@50. This enhancement stemmed from the model's ability to synthesize physically plausible GPR images with multi-scale defect variations, effectively expanding the training data. Subsequent integration of the MCFF module into the neck network achieved +3.5% Precision and +0.9% mAP@50 through multi-scale feature aggregation. MCFF captured cross-resolution defect signatures while suppressing background noise. Incorporation of the GAM into the backbone network refined feature discriminability, yielding incremental gained of +3.3% Precision compared to the original structure. GAM's channel-spatial attention mechanism selectively amplified defect-related features. Final optimization leveraged pre-trained weights from the MS COCO dataset, accelerating convergence while enhancing generalization capacity. This stage delivered additional performance lifts of +1.2% Recall and +0.8% mAP@50, validating the transferability of cross-domain feature representations.

The combination of all enhancements contributed to a notable increase in the algorithm's P, R, and mAP@50, reaching 92.8%, 94.0%, and 96.7%. These outcomes illustrate the efficacy of each enhancement in enhancing the model's overall performance. The MCGA-Net, incorporating data augmentation, enhanced global attention, improved feature extraction, and transfer learning, exhibited excellent object detection capabilities. It was highly suitable for practical applications in underground defect detection.

TABLE VIII
ABLATION EXPERIMENTS

| Data augmentation | MCFF | GAM | MS COCO pre-trained | Precision (%) | Recall (%) | mAP@50 (%) |
|---|---|---|---|---|---|---|
|  |  |  |  | 75.4 | 76.9 | 80.9 |
| √ |  |  |  | 89.1(+13.7) | 92.5(+15.6) | 94.8(+13.9) |
| √ | √ |  |  | 92.6(+17.2) | 92.7(+15.8) | 95.7(+14.8) |
| √ |  | √ |  | 92.4(+17.0) | 92.4(+15.5) | 94.9(+14.0) |
| √ | √ | √ |  | 92.8(+17.4) | 92.8(+15.9) | 95.9(+15.0) |
| √ | √ | √ | √ | **92.8(+17.4)** | **94.0(+17.1)** | **96.7(+15.8)** |

*D. Comparison experiments*

To evaluate the effectiveness of the proposed algorithm, a series of comparative experiments were conducted using classical algorithm and various iterations of the YOLO algorithm. The tested algorithms included Faster R-CNN, SSD, YOLOv3 Tiny, YOLOv5, YOLOv6, YOLOv8, YOLOv8+MCFF, and MCGA-Net. To ensure consistency in training conditions, all models were trained on the same dataset with identical parameter settings over 160 epochs.

Table X summarizes the results of these comparative experiments, presenting performance metrics for each model in terms of Precision, Recall, and mAP@50. MCGA-Net achieved the highest scores of 92.8% Precision, 92.8% Recall, and 95.9% mAP@50. These results underscore the efficacy of the model modifications introduced in this study. Furthermore, the mAP@50 comparison revealed that single-stage algorithms (SSD, YOLO series) exhibited faster convergence than two-stage algorithms (Faster R-CNN) under the 160-epoch training regime.

Fig. 13 provides a visual comparison of detection performance across algorithms. While Faster R-CNN displayed high bounding box confidence scores, it suffered from higher false positive rate compared to YOLO-based models. Although various iterations of the YOLO algorithm accurately localized defect regions, their confidence scores showed negligible improvement. In contrast, MCGA-Net achieved the highest detection accuracy for diverse hidden defect types, highlighting the impact of its targeted enhancements on interpreting complex GPR features.

To validate MCGA-Net's robustness under noisy conditions, Gaussian noise (μ=0, σ=25) was injected into real GPR images. As shown in Fig. 14, YOLOv8 experienced decline in confidence scores, whereas MCGA-Net maintained stable detection accuracy and even improved confidence, attributed to its noise-resistant MCFF and GAM modules.

Additionally, MCGA-Net demonstrated significant improvements in detecting weak signals and small targets. Fig. 15 compares its performance against baseline models for low-amplitude cavities, small crack and concave. The integration of GAM and MCFF modules increased confidence scores by about 10%, respectively, with MCGA-Net achieving optimal results. This enhancement is critical for identifying subtle subsurface anomalies in cluttered GPR data.

TABLE IX
COMPARISON EXPERIMENTS

| Model | Precision (%) | Recall (%) | mAP@50 (%) |
|---|---|---|---|
| Faster R-CNN | 37.4 | 83.3 | 73.5 |
| SSD | **95.5** | 75.0 | 92.4 |
| YOLOv3- tiny | 95.2 | 84.7 | 93.1 |
| YOLOv5 | 91.4 | 90.4 | 94.3 |
| YOLOv6 | 88.7 | 92.4 | 95.1 |
| YOLOv8 | 89.1 | 92.5 | 94.8 |
| YOLOv8+MCFF | 92.6 | 92.7 | 95.7 |
| MCGA-Net | 92.8 | **92.8** | **95.9** |

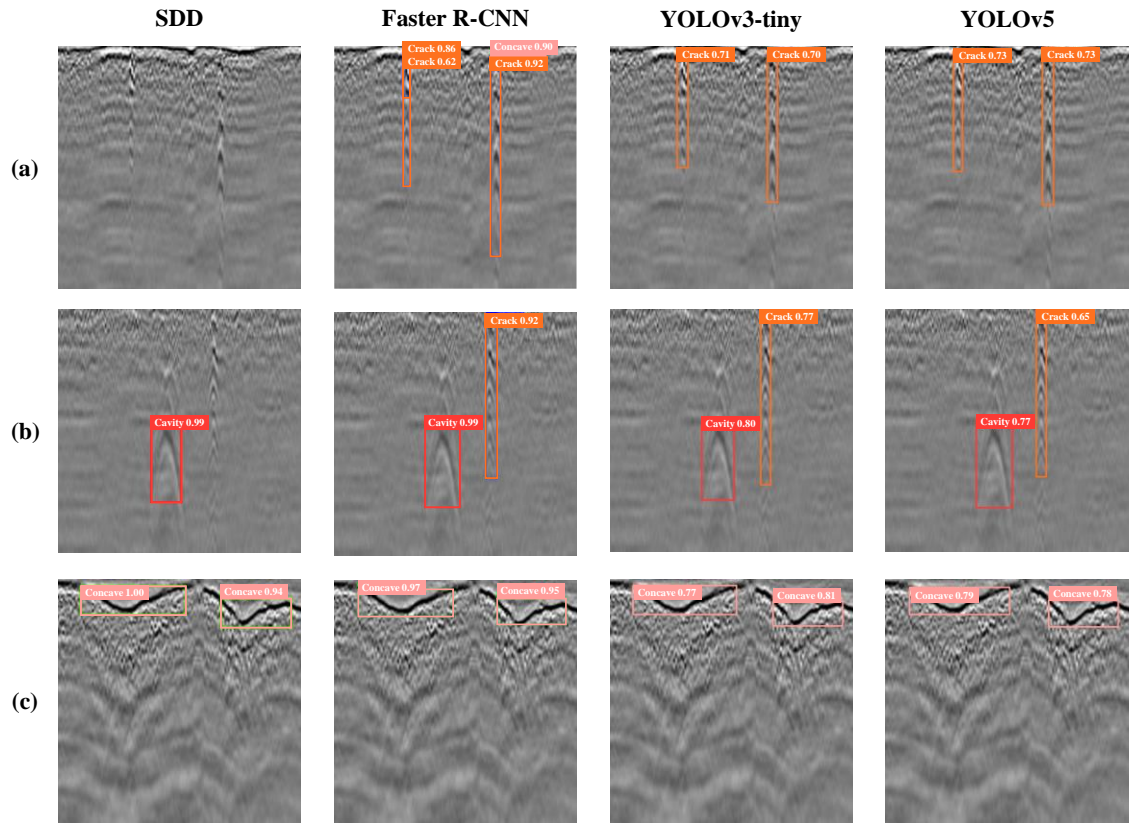

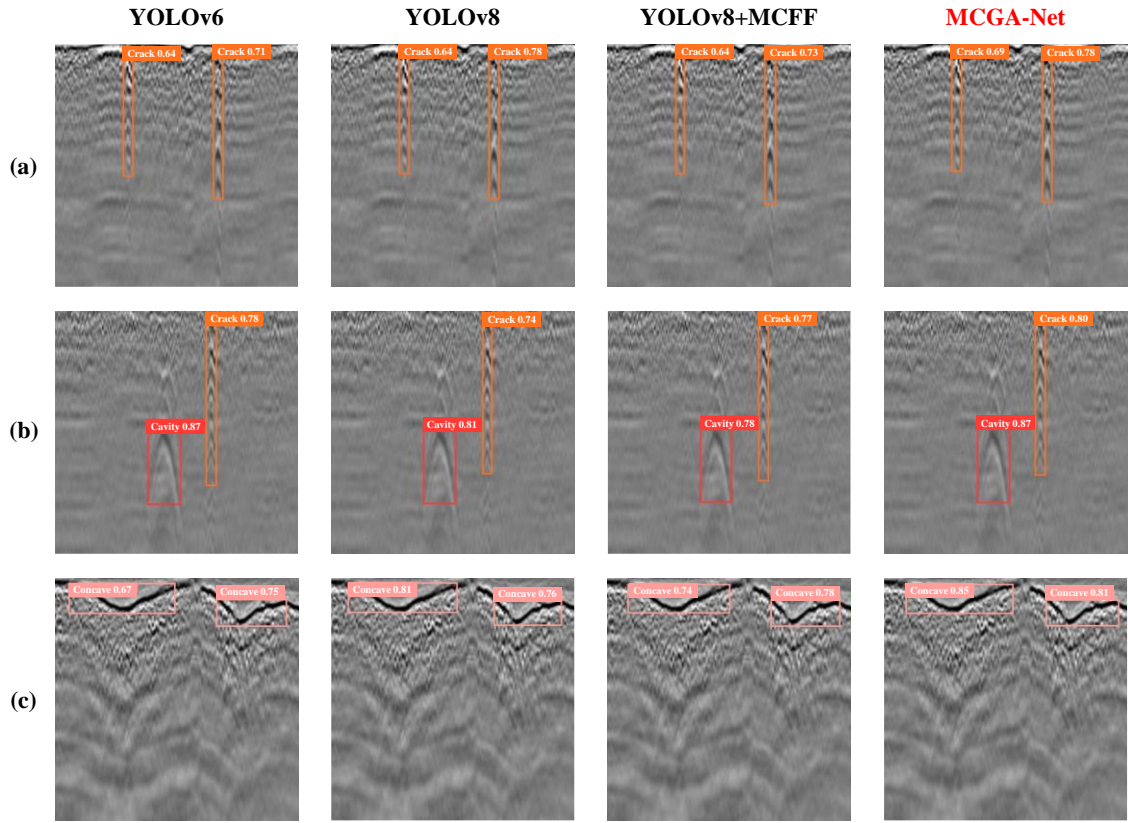

**Fig. 13.** Detection comparison results of different models

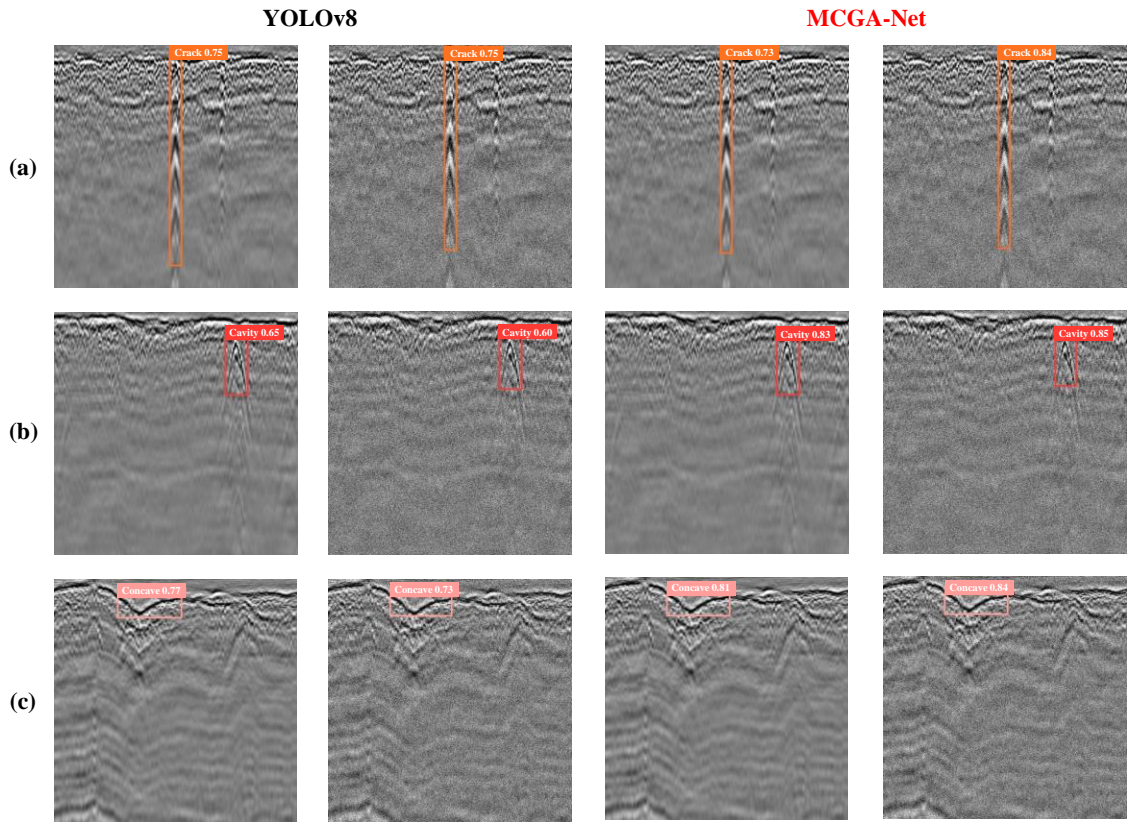

**Fig. 14.** Detection comparison results of before and after increasing interference noise

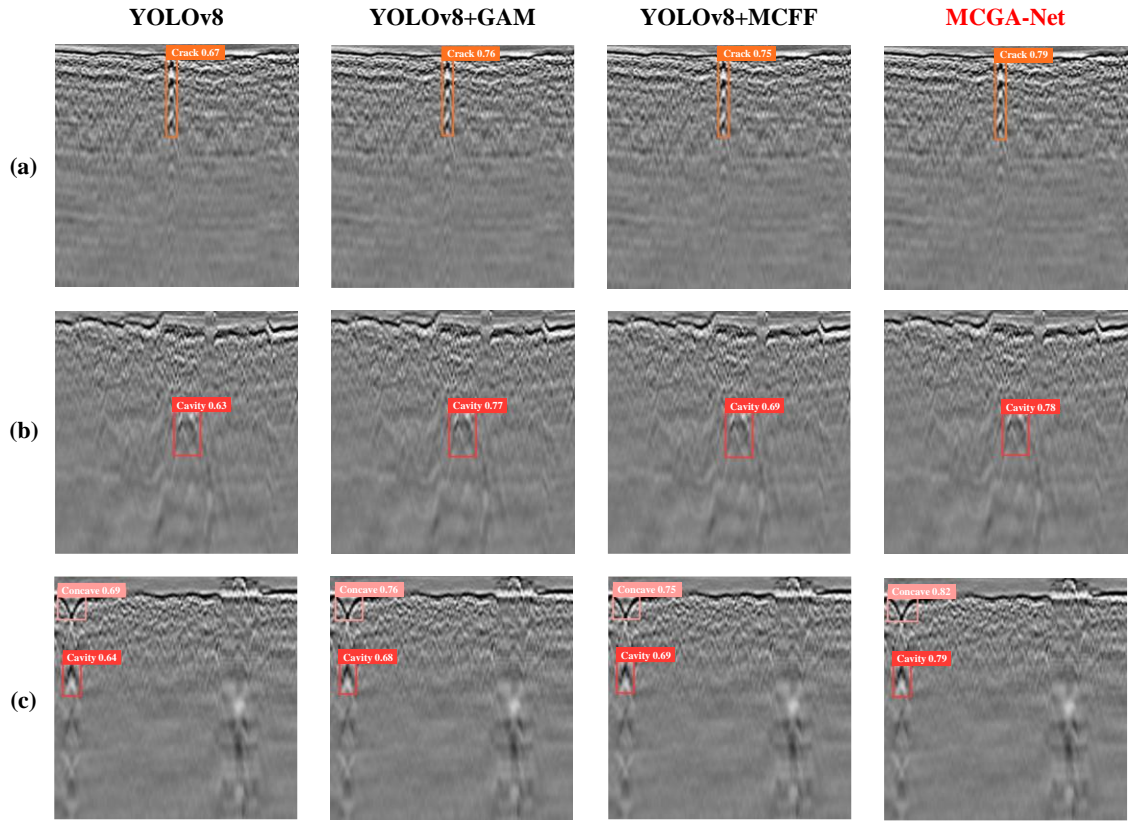

**Fig. 15.** Detection comparison results of weak signals and small targets

## VI. Conclusion

This study proposes a novel multi-category road hidden defect recognition framework, which synergistically enhances feature extraction quality and cross-dimensional feature interaction for intelligent GPR-based hidden defect recognition. The key conclusions are as follows:

1. The DCGAN framework outperformed conventional GANs in generating high-fidelity GPR images. As the training epochs increased, FID significantly decreased, reaching 31.26 at 4000 epochs. Energy gradient measures indicated that the complexity of the generated image is comparable to that of the original image. Compared to baseline models trained on non-augmented data, Precision (+13.7%), Recall (+12.7), and mAP@50 (+11.2%) were all higher. This addressed the scarcity of real-world GPR data while preserving subsurface defect morphology under complex backgrounds.

2. The integration of MCFF and GAM module into YOLOv8 significantly improved defect detection. MCFF enabled hierarchical fusion of multi-scale features. Moreover, GAM amplified defect-related spatial-channel interactions. It achieved Precision (+3.7%), Recall (+0.3%), and mAP@50 (+1.1%) improvements over original YOLOv8.

3. The framework demonstrated robust performance in GPR detection for hidden defects on actual roads. It achieved 96.7% mAP@50 for cavities, concaves, and cracks under variable subsurface conditions. In addition, it reduced the dependence of DL on the amount of real GPR data, improved defect recognition performance, and promoted the applicability of DL in actual road hidden defect inspection.

Future work will pay more attention to identifying and classifying hidden defects in roads. For example, moisture, bonding failure between interlayers, and loose need to be accurately identified and classified. In addition, using multi-view GPR images can more accurately estimate the size, shape, and direction of defects. Meanwhile, in practical applications, relying on GPR data may not be sufficient to capture all relevant structural features. Future research can investigate the integration of multimodal data, such as combining GPR with LiDAR, to improve the accuracy and reliability of overall detection.

### Acknowledgment

This research was funded by the National Key Research and Development Program of China [grant number 2023YFB2603500].

### References

[1] S. W. Li, X. Y. Gu, X. R. Xu, D. W. Xu, T. J. Zhang, Z. Liu, and Q. Dong, "Detection of concealed cracks from ground penetrating radar images based on deep learning algorithm," *Construction and Building Materials*, vol. 273, Mar, 2021.

[2] J. H. Xie, F. J. Niu, W. J. Su, Y. H. Huang, and G. G. Liu, "Identifying airport runway pavement diseases using complex signal analysis in GPR post-processing," *Journal of Applied Geophysics,* vol. 192, Sep, 2021.

[3] T. X. H. Luo, and W. W. L. Lai, "GPR pattern recognition of shallow subsurface air voids," *Tunnelling and Underground Space Technology,* vol. 99, May, 2020.

[4] K. Wang, J. W. Zhang, G. Q. Gao, J. L. Qiu, Y. J. Zhong, C. X. Guo, W. C. Zhao, K. J. Tang, and X. L. Su, "Causes, Risk Analysis, and Countermeasures of Urban Road Collapse in China from 2019 to 2020," *Journal of Performance of Constructed Facilities,* vol. 36, no. 6, Dec, 2022.

[5] H. Yao, Z. J. Xu, Y. Hou, Q. Dong, P. F. Liu, Z. J. Ye, X. Pei, M. Oeser, L. B. Wang, and D. W. Wang, "Advanced industrial informatics towards smart, safe and sustainable roads: A state of the art," *Journal of Traffic and Transportation Engineering-English Edition,* vol. 10, no. 2, pp. 143-158, Apr, 2023.

[6] F. Li, F. Yang, R. Yan, X. Qiao, H. Xing, and Y. Li, "Study on Significance Enhancement Algorithm of Abnormal Features of Urban Road Ground Penetrating Radar Images," *Remote Sensing,* vol. 14, no. 7, Apr, 2022.

[7] V. Marecos, S. Fontul, M. D. Antunes, and M. Solla, "Evaluation of a highway pavement using non-destructive tests: Falling Weight Deflectometer and Ground Penetrating Radar," *Construction and Building Materials,* vol. 154, pp. 1164-1172, Nov, 2017.

[8] G. L. Yuan, A. L. Che, and S. K. Feng, "Evaluation method for the physical parameter evolutions of highway subgrade soil using electrical measurements," *Construction and Building Materials,* vol. 231, Jan, 2020.

[9] S. Sivagnanasuntharam, A. Sountharajah, and J. Kodikara, "In-situ spot test measurements and ICMVs for asphalt pavement: lack of correlations and the effect of underlying support," *International Journal of Pavement Engineering,* vol. 24, no. 1, Dec, 2023.

[10] K. Namgyu, K. Sehoon, A. Y. Kyu, and L. J. Jae, "A novel 3D GPR image arrangement for deep learning-based underground object classification," *International Journal of Pavement Engineering,* vol. 22, no. 6, 2021.

[11] J. Zhang, X. Yang, W. Li, S. Zhang, and Y. Jia, "Automatic detection of moisture damages in asphalt pavements from GPR data with deep CNN and IRS method," *Automation in Construction,* vol. 113, 2020.

[12] N. Kim, K. Kim, Y. K. An, H. J. Lee, and J. J. Lee, "Deep learning-based underground object detection for urban road pavement," *International Journal of Pavement Engineering,* vol. 21, no. 13, pp. 1638-1650, Nov, 2020.

[13] H. Q. Xiong, J. Li, Z. L. Li, and Z. Y. Zhang, "GPR-GAN: A Ground-Penetrating Radar Data Generative Adversarial Network," *Ieee Transactions on Geoscience and Remote Sensing,* vol. 62, 2024.

[14] H. Qin, D. H. Zhang, Y. Tang, and Y. Z. Wang, "Automatic recognition of tunnel lining elements from GPR images using deep convolutional networks with data augmentation," *Automation in Construction,* vol. 130, Oct, 2021.

[15] B. Wang, K. P. Li, S. R. Wu, and P. Y. Chen, "GPR B-Scan Image Augmentation via GAN With Multiscale Discrimination Strategy," *Ieee Transactions on Geoscience and Remote Sensing,* vol. 62, 2024.

[16] F. J. Niu, Y. H. Huang, P. F. He, W. J. Su, C. L. Jiao, and L. Ren, "Intelligent recognition of ground penetrating radar images in urban road detection: a deep learning approach," *Journal of Civil Structural Health Monitoring*, 2024 Jul, 2024.

[17] L. J. Li, L. Yang, Z. Y. Hao, X. L. Sun, and G. F. Chen, "Road sub-surface defect detection based on gprMax forward simulation-sample generation and Swin Transformer-YOLOX," *Frontiers of Structural and Civil Engineering,* vol. 18, no. 3, pp. 334-349, Mar, 2024.

[18] N. J. Zhou, J. M. Tang, W. X. Li, Z. Y. Huang, and X. N. Zhang, "Application of clustering algorithms to void recognition by 3D ground penetrating radar," *Frontiers in Materials,* vol. 10, Sep, 2023.

[19] Z. Tong, J. Gao, and D. D. Yuan, "Advances of deep learning applications in ground-penetrating radar: A survey," *Construction and Building Materials,* vol. 258, Oct, 2020.

[20] M. Rasol, J. C. Pais, V. Perez-Gracia, M. Solla, F. M. Fernandes, S. Fontul, D. Ayala-Cabrera, F. Schmidt, and H. Assadollahi, "GPR monitoring for road transport infrastructure: A systematic review and machine learning insights," *Construction and Building Materials,* vol. 324, Mar, 2022.

[21] H. Liu, C. X. Lin, J. Cui, L. S. Fan, X. Y. Xie, and B. F. Spencer, "Detection and localization of rebar in concrete by deep learning using ground penetrating radar," *Automation in Construction,* vol. 118, Oct, 2020.

[22] J. Zhang, H. W. Li, X. K. Yang, Z. Cheng, P. X. W. Zou, J. Gong, and M. Ye, "A novel moisture damage detection method for asphalt pavement from GPR signal with CWT and CNN," *Ndt & E International,* vol. 145, Jul, 2024.

[23] H. Hu, H. Fang, N. Wang, D. Ma, J. Dong, B. Li, D. Di, H. Zheng, and J. Wu, "Defects identification and location of underground space for ground penetrating radar based on deep learning," *Tunnelling and Underground Space Technology,* vol. 140, Oct, 2023.

[24] B. Zhang, H. Y. Cheng, Y. H. Zhong, J. Chi, G. Y. Shen, Z. X. Yang, X. L. Li, and S. J. Xu, "Real-Time Detection of Voids in Asphalt Pavement Based on Swin-Transformer-Improved YOLOv5," *Ieee Transactions on Intelligent Transportation Systems,* vol. 25, no. 3, pp. 2615-2626, Mar, 2024.

[25] J. G. Yang, K. G. Ruan, J. Gao, S. G. Yang, and L. C. Zhang, "Pavement Distress Detection Using Three-Dimension Ground Penetrating Radar and Deep Learning," *Applied Sciences-Basel,* vol. 12, no. 11, Jun, 2022.

[26] Z. Liu, S. Q. Wang, X. Y. Gu, D. Y. Wang, Q. Dong, and B. Y. Cui, "Intelligent Assessment of Pavement Structural Conditions: A Novel FeMViT Classification Network for GPR Images," *Ieee Transactions on Intelligent Transportation Systems,* vol. 25, no. 10, pp. 13511-13523, Oct, 2024.

[27] C. Liu, Y. S. Yao, J. Li, J. F. Qian, and L. H. Liu, "Research on lightweight GPR road surface disease image recognition and data expansion algorithm based on YOLO and GAN," *Case Studies in Construction Materials,* vol. 20, Jul, 2024.

[28] X. G. Guo, and N. Wang, "Automated Identification of Pavement Structural Distress Using State-of-the-Art Object Detection Models and Nondestructive Testing," *Journal of Computing in Civil Engineering,* vol. 38, no. 4, Jul, 2024.

[29] N. N. Wang, Z. X. Zhang, H. B. Hu, B. Li, and J. W. Lei, "Underground Defects Detection Based on GPR by Fusing Simple Linear Iterative Clustering Phash (SLIC-Phash) and Convolutional Block Attention Module (CBAM)-YOLOv8," *Ieee Access,* vol. 12, pp. 25888-25905, 2024.

[30] Z. Liu, X. Y. Gu, J. Li, Q. Dong, and J. W. Jiang, "Deep learning-enhanced numerical simulation of ground penetrating radar and image detection of road cracks," *Chinese Journal of Geophysics-Chinese Edition,* vol. 67, no. 6, pp. 2455-2471, Jun, 2024.

[31] A. Joshaghani, and M. Shokrabadi, "Ground penetrating radar (GPR) applications in concrete pavements," *International Journal of Pavement Engineering,* vol. 23, no. 13, pp. 4504-4531, Nov, 2022.

[32] E. Eide, P. A. Valand, and J. Sala, "Ground-Coupled Antenna Array for Step-Frequency GPR." pp. 756-761, 2014.

[33] K. Yan, Z. H. Zhang, and X. L. Xu, "Improved Tucker decomposition algorithm for noise suppression of 3D GPR data in road detection," *Near Surface Geophysics,* vol. 21, no. 2, pp. 138-151, Apr, 2023.

[34] I. Goodfellow, J. Pouget-Abadie, M. Mirza, B. Xu, D. Warde-Farley, S. Ozair, A. Courville, and Y. Bengio, "Generative Adversarial Networks," *Communications of the Acm,* vol. 63, no. 11, pp. 139-144, Nov, 2020.

[35] A. Radford, L. Metz, and S. Chintala, "Unsupervised Representation Learning with Deep Convolutional Generative Adversarial Networks," *Arxiv*, Jan 07, 2016.

[36] M. Brazell, N. Li, C. Navasca, and C. Tamon, "SOLVING MULTILINEAR SYSTEMS VIA TENSOR INVERSION," *Siam Journal on Matrix Analysis and Applications,* vol. 34, no. 2, pp. 542-570, 2013.

[37] J. Hu, L. Shen, G. Sun, and Ieee, "Squeeze-and-Excitation Networks," *IEEE Conference on Computer Vision and Pattern Recognition.* pp. 7132-7141, 2018.

[38] S. H. Woo, J. Park, J. Y. Lee, and I. S. Kweon, "CBAM: Convolutional Block Attention Module," *Lecture Notes in Computer Science.* pp. 3-19, 2018.

[39] J. Redmon, S. Divvala, R. Girshick, A. Farhadi, and Ieee, "You Only Look Once: Unified, Real-Time Object Detection," *IEEE Conference on Computer Vision and Pattern Recognition.* pp. 779-788, 2016.

[40] T. Y. Lin, M. Maire, S. Belongie, J. Hays, P. Perona, D. Ramanan, P. Dollár, and C. L. Zitnick, "Microsoft COCO: Common Objects in Context," *Lecture Notes in Computer Science.* pp. 740-755, 2014.

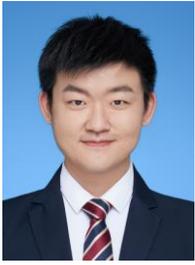
**Haotian Lv** received the B.S. degree in Civil Engineering from Shandong University, Jinan, China, in 2020. He has been a Ph.D. candidate in the School of Transportation Science and Engineering, Harbin Institute of Technology, Harbin, China since 2020. His research interests include non-destructive testing, Ground Penetrating Radar (GPR) and Deep Learning (DL) methods in GPR image recognition.

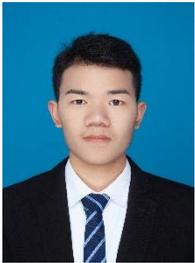
**Yuhui Zhang** received the B.S. degree in Road, Bridge and Cross River Engineering from Zhengzhou University, Henan, China, in 2023. He is currently working toward the M.S. degree in Traffic and Transportation Engineering with the School of Transportation Science and Engineering, Harbin Institute of Technology, Harbin, China. His research interests include non-destructive testing, ground penetrating radar (GPR) and data fusion.

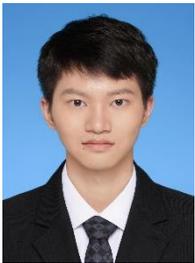
**Jiangbo Dai** received the B. S. degree in Road and Bridge Engineering from Harbin Institute of Technology, Harbin, China, in 2024. He has been a Ph.D. candidate in the School of Transportation Science and Engineering, Harbin Institute of Technology, Harbin, China since 2024. His research interests include ground penetrating radar (GPR) detection and processing of its data.

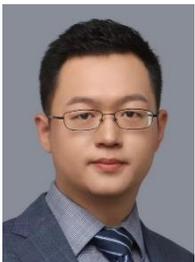
**Hanli Wu** received his B.S. and M.S. degree in Civil Engineering in Hunan University in 2019. In 2018, he joined the Department of Civil, Architectural, and Environmental Engineering at Missouri S&T and earned his Ph.D. degree in 2022. After that, he continued his research at Missouri S&T as a postdoc research associate and lecturer. He joined Harbin Institute of Technology as an associate professor in 2023. His research interests and expertise focus primarily on frozen ground engineering, Multiphysics modeling of climate-soil-structure interaction, fluid-thermal-structure coupling, high-performance computing, structural health monitoring, and bridge inspection. He was actively involved in professional civil engineering organizations, such as the American Society of Civil Engineers (ASCE), the International Society for Structural Health Monitoring of Intelligent Infrastructure (ISHMII), and the International Society of Chinese Infrastructure Professionals (IACIP).

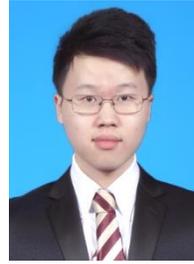
**Jiaji Wang** received his B.S. and Ph.D. degree in Civil Engineering in Tsinghua University in 2014 and 2019. He joined the Department of Civil Engineering at the University of Hong Kong as Assistant Professor in 2023. His research is focused on Data-driven and physics-informed operator learning for solving partial differential equations and AI-based structural health monitoring. He has been awarded the outstanding award of China Steel Construction Society in 2019 and published 39 journal papers as first or corresponding author in leading journals of structural engineering and computational mechanics with more than 910 citations and H-index of 18.

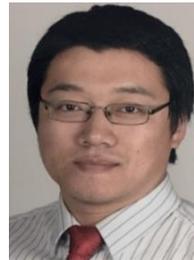
**Dawei Wang** (Senior Member, lEEE) received the B.S. in civil engineering in Tsinghua University in 2003, M.S. and Ph.D. in roadway engineering in RWTH Aachen University in 2007 and 2011, respectively. In 2017, he was granted Habilitation based on the research and effort he has contributed to the highway engineering in Germany. His research interests and expertise focus primarily on asphalt pavement skid resistance, multi-scale characterization of the asphalt pavement mechanical behavior and functional pavement theory and technology. So far, he has directed more than 24 scientific research projects. Over the last years, he has published nearly 200 academic publications, including more than 140 SCI indexed papers. He also serves on the editorial boards of many international academic journals.